\def\year{2022}\relax
\documentclass[letterpaper]{article} % DO NOT CHANGE THIS
\usepackage{aaai22}  % DO NOT CHANGE THIS
\usepackage{times}  % DO NOT CHANGE THIS
\usepackage{helvet}  % DO NOT CHANGE THIS
\usepackage{courier}  % DO NOT CHANGE THIS
\usepackage[hyphens]{url}  % DO NOT CHANGE THIS
\usepackage{graphicx} % DO NOT CHANGE THIS
\def\UrlFont{\rm}  % DO NOT CHANGE THIS
\usepackage{natbib}  % DO NOT CHANGE THIS AND DO NOT ADD ANY OPTIONS TO IT
\usepackage{caption} % DO NOT CHANGE THIS AND DO NOT ADD ANY OPTIONS TO IT
\DeclareCaptionStyle{ruled}{labelfont=normalfont,labelsep=colon,strut=off} % DO NOT CHANGE THIS
\usepackage{algorithm}
\usepackage[switch]{lineno}
\usepackage{algorithmic}
\usepackage{cite}
\usepackage{amsmath,amssymb,amsfonts,amsthm}
\usepackage{textcomp}
\usepackage{xcolor}
\usepackage{kotex}
\usepackage{todonotes}
\usepackage{subcaption}
\usepackage{booktabs}
\usepackage{multirow}
\usepackage{pgfplots}
\usepackage{nicefrac}
\usepackage[algo2e,ruled,vlined]{algorithm2e}
\usepackage{dsfont}
\usepackage{framed}
\usepackage{thmtools}
\usepackage{tabularx}
\usepackage{pgfplots}
\usepackage{balance}
\usepackage{color, colortbl}

\newcommand{\ours}{\textsf{DAS}}
\usepackage{newfloat}
\usepackage{listings}
\usepackage[hang,flushmargin]{footmisc}
\floatstyle{ruled}
\newfloat{listing}{tb}{lst}{}
\floatname{listing}{Listing}
\nocopyright
\title{
Knowledge Sharing via Domain Adaptation in Customs Fraud Detection 
}
\author{
    Sungwon Park\textsuperscript{\rm 1},
    Sundong Kim\textsuperscript{\rm 2}\textsuperscript{\dag},
    Meeyoung Cha\textsuperscript{\rm 2,1}\textsuperscript{\dag}
}
\affiliations{
    \textsuperscript{\rm 1}School of Computing, Korea Advanced Institute of Science and Technology\\
    \textsuperscript{\rm 2}Data Science Group, Institute for Basic Science\\
}

\usepackage{bibentry}
\begin{document}
% \linenumbers
\maketitle
\begingroup\renewcommand\thefootnote{\dag}
\noindent \footnotetext{Corresponding authors}
\endgroup

\begin{abstract}
Knowledge of the changing traffic is critical in risk management. Customs offices worldwide have traditionally relied on local resources to accumulate knowledge and detect tax fraud. This naturally poses countries with weak infrastructure to become tax havens of potentially illicit trades. The current paper proposes \ours{}, a memory bank platform to facilitate knowledge sharing across multi-national customs administrations to support each other. We propose a domain adaptation method to share transferable knowledge of frauds as prototypes while safeguarding the local trade information. Data encompassing over 8 million import declarations have been used to test the feasibility of this new system, which shows that participating countries may benefit up to 2--11 times in fraud detection with the help of shared knowledge. We discuss implications for substantial tax revenue potential and strengthened policy against illicit trades. 
\end{abstract}

\section{Introduction}

\noindent
Customs administrations manage an astronomical amount of trades. Amongst their tasks is the risk management and detection of irregularities and illicit consignments from import declarations. These tasks are critical as import tariffs account for a substantial proportion of the total tax revenue. The detection process has traditionally used rule-based algorithms, which is gradually changing to machine learning algorithms~\cite{mikuriya2021wcj}. The World Customs Organization (WCO) has been leading such data initiatives by assisting customs offices in 175 countries with their digital transformation process~\cite{weerth2009}.

\begin{figure}[t!]
    \centerline{
    \includegraphics[width=0.9\columnwidth]{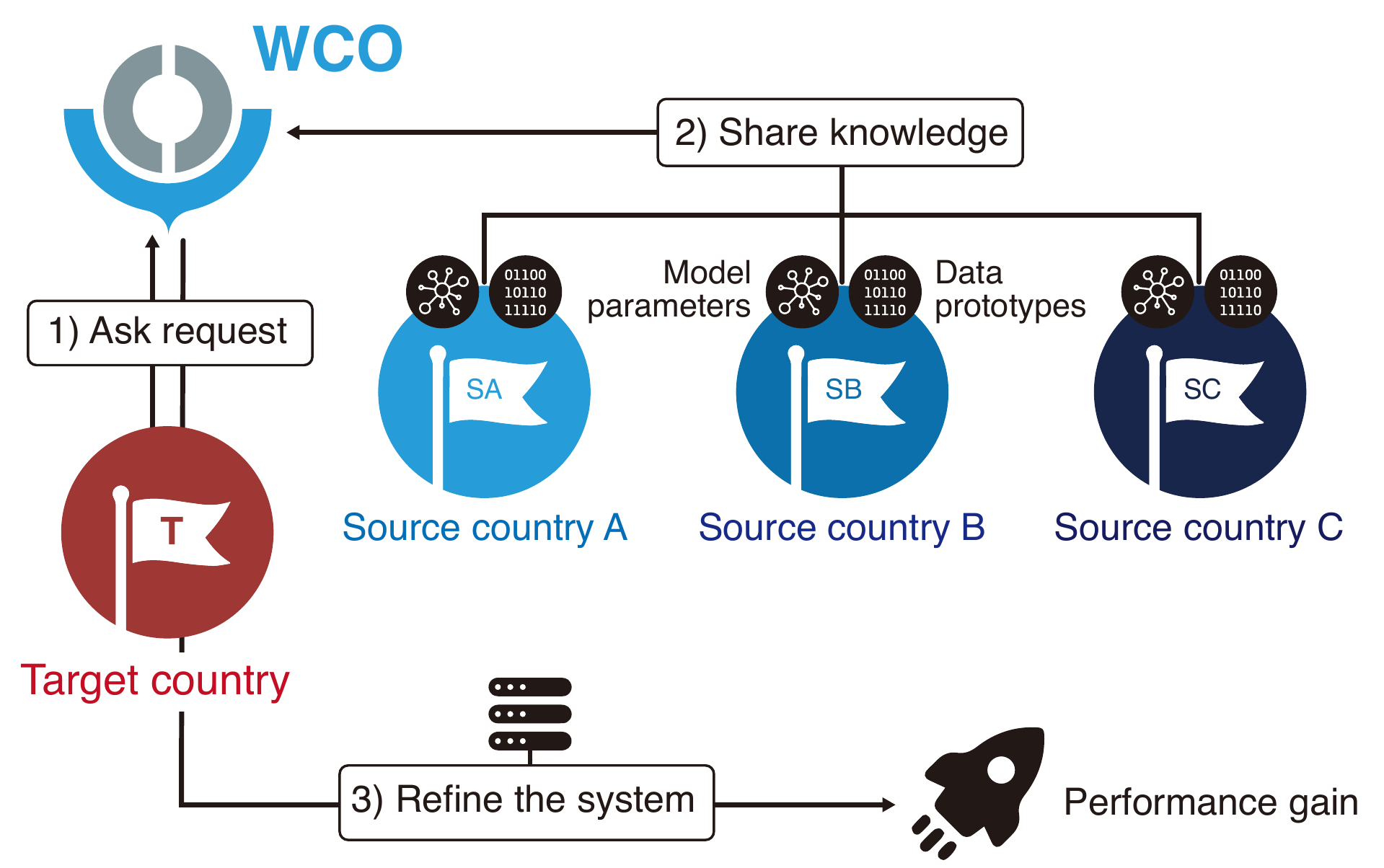}}
    \caption{
    A theoretical model framework describing the workflow to facilitate knowledge sharing across multiple countries.
    %Workflow to facilitate knowledge sharing across members of the World Customs Organization.
    } 
    \label{fig:intro}
\end{figure}

Several deep learning algorithms have been proposed and tested at a regional level~\cite{vanhoeyveld2020belgian, kim2020date}. However, these advanced models only benefit customs offices that have the capacity to build deep learning models and train them. Many developing and low-income countries do not have such data infrastructure. One method to bootstrap model learning is to utilize shared logs of frauds across custom offices within geographical proximity, as illicit patterns are likely similar in those regions. Sharing knowledge can facilitate data initiative and strengthen policy against illegal trade activities. However, data sharing has not been considered due to privacy concerns, as trade information entails critical industry and business relationships.  
 
This research proposes a first-of-a-kind collaborative customs fraud detection system that enables information sharing across regional boundaries. We use techniques in domain adaptation and propose a new data prototyping method to extract transferable knowledge from import declarations. Figure~\ref{fig:intro} illustrates the workflow. Data donors or source countries share logs of speculative trades, for instance, 1,000 import declarations after removing identifiable information like importer names. Information about the vast majority of normal trades is not shared. Furthermore, instead of the raw data, its embedding is shared in the form of prototypes. At the receiving end, target countries access the accumulated knowledge of fraud patterns from the memory bank and refine their local knowledge via soft attention. Target countries that typically lack infrastructure will likely observe new fraud transactions thanks to the model. This information will loop back to the central memory bank to strengthen the combined knowledge and be shared with all contributing members. \looseness=-1 

Our memory bank design is called \ours{} (Domain Adaptation for Sharing Knowledge in Customs), and it uses contrastive learning and clustering to extract meaningful fraud characteristics. Experiments based on multi-year, million-scale import declarations show that all participating countries---including data contributors and recipients---can benefit from the transferred knowledge with an estimated 2 to 11 times increment in the total tax raised from fraud detection, far extending the capability to manage risk in the studied region.

In practice, \ours{} can be adopted among countries under bilateral and regional trade agreements to accelerate market opening and pursue higher levels of trade freedom. Globally more countries are starting to share tariff rates, relieve trade barriers, and adopt common policies. For countries initiating trade agreements, implementing this system will facilitate active knowledge sharing and derive tangible efforts on strengthening border security.

\section{Related Work}

\paragraph{Customs Fraud Detection Algorithms.} 

Earlier efforts on customs fraud detection utilized rule-based algorithms and random selection algorithms~\cite{hua2006rule}. While some customs offices have adopted machine learning~\cite{sisam2015paper}, many offices in developing countries still report their reliance to rule systems and expert knowledge~\cite{goldberg2009trade}. Recent studies applied off-the-shelf algorithms, including the ensembled SVM in customs fraud detection~\cite{vanhoeyveld2020belgian}. State-of-the-art models, for example, the Dual Attentive Tree-aware Embedding (DATE) model, employ gradient boosting and attentions to generate transaction-level embeddings and provide interpretable decisions~\cite{kim2020date}. Some newer models utilize concept drift to better represent the changing trade patterns over time~\cite{kim2021take, mai2021drift}.

\paragraph{Domain Adaptation Techniques.} 

Domain adaptation aims to learn universal representations that are domain invariant. Representative techniques include latent distribution alignment between the source and target domains~\cite{tzeng2017adversarial, hoffman2018cycada, long2017deep}. Contrastive learning is used to extract discriminative features between classes~\cite{kang2019contrastive, thota2021contrastive}, and the memory module is used to augment target features using incremental information~\cite{asghar2018progressive, zheng2019unsupervised, liu2020open}. A long-standing problem in domain adaptation is negative transfer, which refers to the abnormal scenarios when the source domain data causes reduced learning performance in the target domain due to a large discrepancy in data distributions~\cite{wang2019characterizing, zhang2020ntsurvey}. Regularization and adaptive source selection methods have been proposed to mitigate this problem~\cite{liu2020open, abuduweili2021adaptive}. Most domain adaptation techniques assume that the source and the target data can be accessed concurrently, which may not be practical for customs under multi-national administrations. 

% reference 추가 (% reference 길이는 1페이지 꽉 차게 맞췄습니다.)
\nocite{saito2018maximum, li2019joint, tang2020discriminative, hoffman2018cycada}
\nocite{mirror2015, Zhou2019WCO}

% : adversarial discrimination, tzeng2017adversarial}, cycle consistency~\cite{hoffman2018cycada}, and joint maximum mean discrepancy~\cite{long2017deep}.

%% 너무 길어져서, 안 쓰이는 부분은 지웠음
% Modern models even provide interpretable decisions that can be validated and checked by human customs officers. 
% Matching task-specific feature distributions between classes in addition to reduce domain discrepancy is critical~\cite{saito2018maximum, li2019joint, tang2020discriminative}.
% : adversarial discrimination, tzeng2017adversarial}, cycle consistency~\cite{hoffman2018cycada}, and joint maximum mean discrepancy~\cite{long2017deep}.
% \input{3_dataset.tex}
\section{Method}

%To enable knowledge sharing across customs offices, we consider the following problem statement.

\subsection{Problem Statement}
%  Let denote import transaction instance and  its corresponding fraud label in custom transactions dataset as $(\mathbf{x}, y) \in \mathcal{D}$.
% A fraud detection system $f$ determines whether an import transaction $\mathbf{x}$ is fraudulent, guides customs officers to inspect the transaction. We assume that there are target country $t$, which has limited logs  of import transactions dataset $\mathcal{D}_{t}$ and the source country $s$ is assumed to possess richer logs than the target country $t$ such that $|\mathcal{D}_s| \gg |\mathcal{D}_t|$. 
% Our task is to design a fraud detection system $f(\cdot|\mathcal{D}_{t})$ for a target country's dataset $\mathcal{D}_{t}$, which has limited data logs. We want to find transferable information from $f(\cdot|\mathcal{D}_{s})$ that can contribute to the performance of the fraud detection system at the target country. 

A fraud detection system $f$ determines whether an import declaration is fraudulent based on its transaction instance $\mathbf{x}$ that is a vector describing information like the imported product and price. $f$ guides customs officers on which transaction to inspect, and customs officers obtain its fraud label $y$ after manual inspection. We denote the inspected import transaction instance and its corresponding fraud label in the custom transactions dataset as $(\mathbf{x}, y) \in \mathcal{D}$.
Our task is to design a fraud detection system $f(\cdot|\mathcal{D}_{t})$ for a target country $t$ that has limited data logs of import transactions $\mathcal{D}_{t}$. Let the inspected transactions of the source country $s$ be $\mathcal{D}_{s}$, then we want to pick out transferable embedded information that will be stored at memory bank $M$. This shared knowledge can be used to improve the fraud detection system at the target country using $f(\cdot|\mathcal{D}_{t}, M)$. The source country $s$ is assumed to possess richer logs than the target country $t$ such that $|\mathcal{D}_s| \gg |\mathcal{D}_t|$.

\subsection{System Overview}

\begin{figure*}[t!]
\centerline{
      \includegraphics[width=0.7\linewidth]{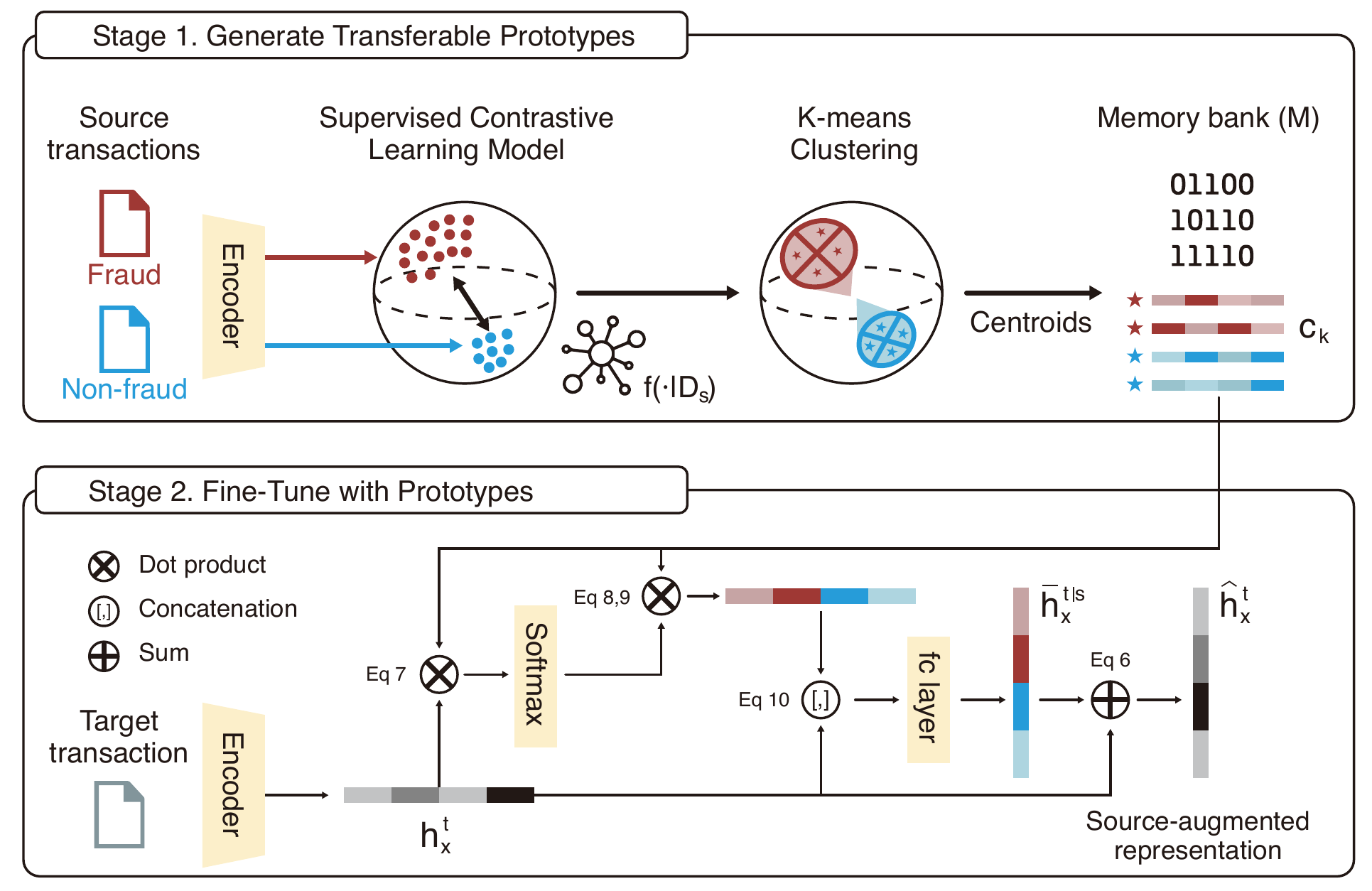}}
      \caption{The overall pipeline depicts how the knowledge shared by the source country is added to the transferable memory bank in Stage 1. The subtle difference between the fraud and non-fraud trades is learned from inspecting speculative logs via contrastive learning. Then, the target country can refine its trade representation in Stage 2. The target country will maintain a better fraud detection system with knowledge-enhanced trade representation.
      %\ryan{model architecture 모델의 아키텍쳐 는 유지하면 임베딩 방법을 개선했다.}
      } 
\label{fig:model}

\end{figure*}
The proposed system employs multiple strategies to ensure safety in data transferability. First is the sampling of abnormal, fraud-like transactions to construct the source dataset $\mathcal{D}_{s}$. Note that most of the ``normal'' import declarations contain critical trading partners and industry information. In contrast, fraud-suspected trades make up a small volume, which the algorithm utilizes. Second, the domain invariant feature of the HS code is enforced, which is a worldwide item notation convention. Additionally, critical information is anonymized or removed, such as country names, importers, declarants, and detailed descriptions of goods. Lastly, we regulate direct data sharing across domains. Here, the knowledge is shared in the form of model parameters and prototypes, representing compact information for a group of semantically similar transactions. These shared prototypes are combined with the local transaction logs to fine-tune the fraud detection model at the target country. Figure~\ref{fig:model} depicts the pipeline of the proposed system in two stages:

%   we considers multiple concepts from existing literature, including contrastive learning~\cite{khosla2020supervised}, prototypical memory bank ~\cite{pcl2021iclr}, and domain adaptation~\cite{tzeng2017adversarial}. 

%% \sd{The major assumption of the transfer learning is to increase the performance of the target task, without sharing its original dataset. I think we can tone down the privacy-preserving part.}
% To preserve privacy of the source data, it is critical to share information that cannot reveal information about the source. 

\begin{itemize}
\item  \textbf{Stage 1} displays the method through which the source country $s$ shares compressed knowledge to the memory bank $M$. $s$ will pretrain a network $f(\cdot|\mathcal{D}_{s})$ using contrastive learning to extract discriminative features from its fraud-suspected logs. The prototype set $\mathcal{C}_s$ is the resulting transferable knowledge that is stored at $M$. 

\item  \textbf{Stage 2} describes the method through which the target country $t$ refines its detection model with the transferred knowledge. Given a pretrained network $f(\cdot|\mathcal{D}_{s})$ and the memory bank $M$ in Stage 1, the data representation at $t$ is augmented using compressed knowledge $\mathcal{C}_s$ in $M$ to fine-tune the network $f(\cdot|\mathcal{D}_{t})$.
\end{itemize}

% \sd{Dummy algorithm}
% \input{algorithms/algo3.tex}

\subsection{%Base Component: 
Domain Invariant Feature using HS code}

% \item  \textbf{Domain invariant network architecture.} 
% Learn domain invariant representation by improving pairwise correlations between item feature and transaction feature.

\begin{figure}[h!]
    \centerline{
    \includegraphics[width=\columnwidth]{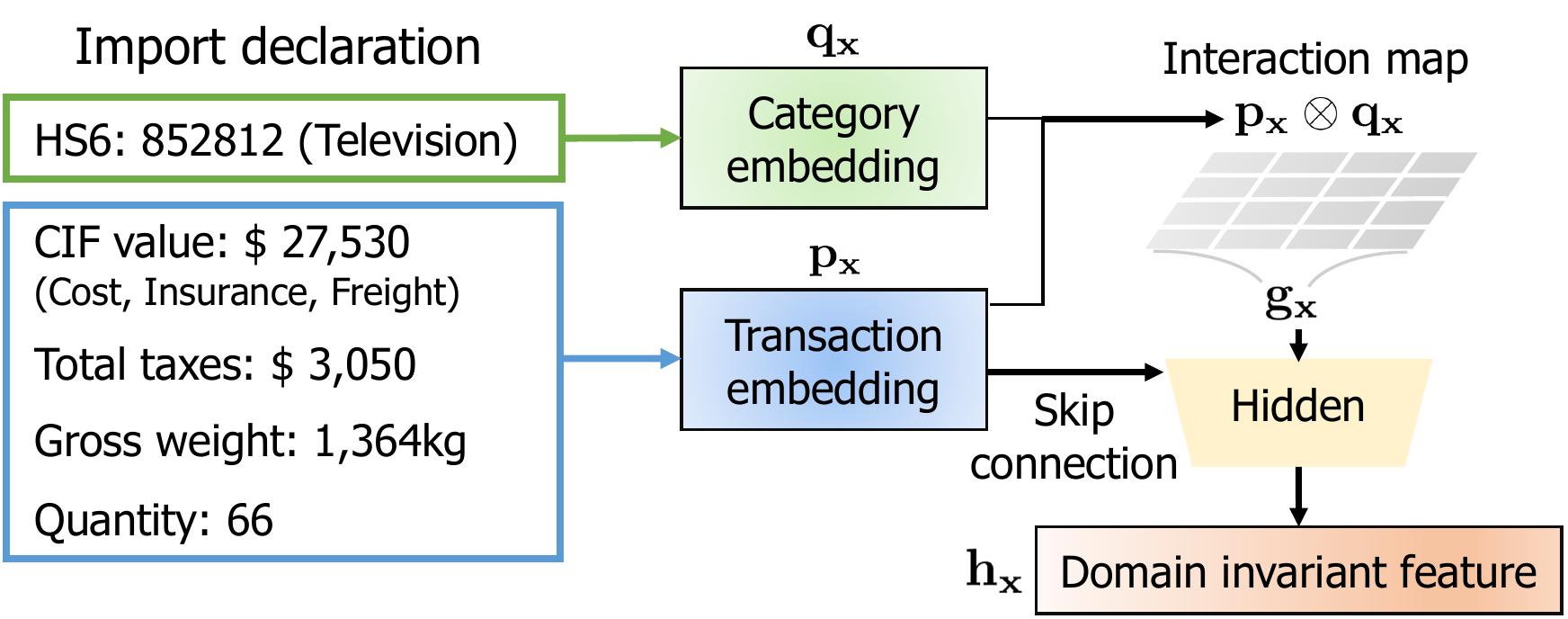}}
    \caption{The domain invariant encoder embeds import declarations by increasing the usability across the country.} 
    \label{fig:encoder}
\end{figure}

A transaction encoder is used in both stages to embed import declarations by enforcing domain invariant information. This encoder augments each transaction log to meet international standards like the HS6 code, i.e., the six-digit category code of goods. Figure~\ref{fig:encoder} describes the design of this new encoder. \looseness=-1

%The encoder is used by both the source and target countries.

% \mc{=what do you mean immutable? also HS6 code can be characteristics of trade secrets=} 
%which is \footnote{Stipulated by the WCO~\cite{goldberg2009trade}, every country uses the same HS6 code for a particular item.} 
% This process \mc{helps knowledge transfer =not intuitive here. maybe say we will demonstrate how this embedding this information can help transfer knowledge...=} between two countries since it \mc{softens =what does this mean?=} the embedding partly learned from country-specific information such as importer, declarant, or the customs office processed the transaction.  

Let $\mathbf{x}$ denote input features of a transaction for a single import declaration. Then its embedding of transaction information, $\mathbf{p_x}$, and the embedding of product category information (i.e., HS code), $\mathbf{q_x}$, are obtained by training a fraud detection model.
%  \ryan{qx 는 Date 임베딩을 사용한 것처럼 보인다.}

Any existing detection model can be used. This research uses the state-of-the-art model that is based on a tree-based embedding for interpretable detection~\cite{kim2020date}. The generated embedding $\mathbf{p_x}$ may reveal the generic trade characteristics of that region. Hence, we do not use a simple concatenation of these two embeddings but instead augment $\mathbf{p_x}$ to increase its interaction with $\mathbf{q_x}$ by taking their outer product. We use Eq.~\eqref{eq:outer_product} to model the pairwise correlation between the embedding space of two variables~\cite{he2018outer}:  
% \footnote{One option is to use DATE~\cite{kim2020date}, which learns transaction, importer, and category embedding to detect fraud transactions.}
% \mc{by training DATE~\cite{kim2020date} =can we not mention DATE but say a tree-based model? DATE is one option. We can mention DATE in the experiment. Here instead, write another paragraph to discuss key components of the model.=}. DATE is a recent fraud detection model for customs fraud detection system, which learns transaction, importer, and category embedding to evaluate the transaction's fraudness.\footnote{Notations in \cite{kim2020date}: $\mathbf{p_x} = \mathbf{e}_{f}(u,c,\tilde{\mathbf{S}}), \mathbf{q_x} = \mathbf{q}_c$.}
%The equation below generates a rich interaction map between the two variables, which helps better encode the information compared to simple concatenations: % this line is added to address Brian's comment, but it can be commented 
\begin{equation}
    \mathbf{E} = \mathbf{p_{x}} \otimes \mathbf{q_{x}} = \mathbf{p_{x}}\mathbf{q_{x}}^{T},
    \label{eq:outer_product}
\end{equation} 
% \noindent \textbf{Outer product interact mapping} A naive approach to interact between $\mathbf{p_x}$ and $\mathbf{q_x}$ is concatenation. Many collaborative filtering models, including DATE, use the concatenation method. Despite its effectiveness, this simple concatenation approach has an inherent limitation which only retains the original information in embeddings without modeling any correlation. So we use an outer product to explicitly model the pairwise correlations between the dimensions of the embedding space as follows:
\noindent where $\mathbf{p_{x}}, \mathbf{q_{x}} \in \mathcal{R}^k$ and $\mathbf{E} \in \mathcal{R}^{k\times k}$. We obtain an interaction vector $\mathbf{g_{x}}$ by taking a convolutions on $\mathbf{E}$ to learn higher-order relations: 
\begin{equation}
    \mathbf{g_x} = \text{CNN}(\mathbf{E}).
    \label{eq:cnn}
\end{equation}
The final representation $\mathbf{h_{x}}$ of transaction $\mathbf{x}$ is refined via the residual connection~\cite{he2016cvpr}:
\begin{equation}
    \mathbf{h_{x}} = \text{ReLU}(\phi_{a}([\mathbf{p_{x}}, \mathbf{g_{x}}])), 
    \label{eq:final_hidden}
\end{equation}

\noindent where $\phi_{a}$ is a learnable function.
% and $[\cdot]$ denotes concatenation. 
% $\mathbf{h_{x}}$ is the final embedded knowledge that will be shared
%We will demonstrate how this encoding process help transfer knowledge between two countries in ablation studies. % this line can be commented

\subsection{Stage 1. Generate Transferable Prototypes}
The encoder projects every transaction $\mathbf{x}$ in $\mathcal{D}_{s}$ to the domain-invariant space $\mathbf{h_{x}}$ that is compatible with knowledge transfer. Then the objective of Stage 1 is to learn discriminative features between fraud and non-fraud transactions. We use contrastive learning to pretrain the network $f(\cdot|\mathcal{D}_{s})$ and extract prototype set $\mathcal{C}_{s}$ from the source representation space.

% There may still exist concerns of leaking item-level representations. We compress the knowledge as a prototype, a representative embedding for a group of semantically similar instances, to ensure that the shared knowledge no longer contains individual trade information. 

% An alternative method will be to only share the parameters of the risk management model $f(\cdot|\mathcal{D}_{s})$ learned using the representation $\mathbf{h_{x}}$ of the source data $\mathcal{D}_{s}$. Nonetheless, the source country may be reluctant to disclose the decision boundary of the fraud detection system due to concerns about adversarial attacks \mc{=reference a paper that shows model parameters can be used for attacks=}. 
% \sd{Model parameter도 전달해 주는 걸로 최종 모델을 정해서 이 부분을 줄였어요.}

\paragraph{Pretrain with Supervised Contrastive Learning.} 
The use of contrastive learning ensures that positive instances belonging to the same class (i.e., frauds or non-frauds) are pulled together in normalized embedding space while simultaneously pushing apart data instances from different classes~\cite{kang2019contrastive, han2020mitigating,chen2020simple}.  We assume source countries own enough inspected logs to leverage the label information effectively. We adopt supervised contrastive learning (SCL) loss  to pretrain the network  $f(\cdot|\mathcal{D}_{s})$ in a fully-supervised manner~\cite{khosla2020supervised}. 
% Here, common cross entropy loss may generalize poorly for the target domain because of suboptimal near the learned decision boundary. These interconnections lead hard negative mining which leads non-trivial generalization for target domain~\cite{kang2019contrastive}.
\begin{equation}
\begin{split} 
   \mathcal{L}_{SCL} = &\sum_{i=1}^N - \frac{1}{N_{y_i}-1} \sum_{j=1}^N \mathbf{1}_{i\neq j} \mathbf{1}_{y_i = y_j} \bigg[ \\
&  \log \frac{\exp(\text{sim}(\mathbf{h_{x_i}} , \mathbf{h_{x_j}}) / \tau)}{\sum_{k=1}^N \mathbf{1}_{i \neq k}\exp(\text{sim}(\mathbf{h_{x_i}} \cdot \mathbf{h_{x_k}}) / \tau)} \bigg],
\end{split}
\label{eq:SCL_loss}
\end{equation}
\noindent where N is the batch size, and $\tau$ is the sharpening temperature.

Compared to using a simple cross-entropy loss, SCL loss is known to increase the transfer learning performance by maximizing the discrepancy between separate classes~\cite{kang2019contrastive, khosla2020supervised}. In the experiment, we verify that the pretrained network using SCL loss shows non-trivial performance improvement.
\paragraph{Compress Knowledge by Clustering.}
Vanilla transfer learning (i.e., passing only model parameters) may suffer from the catastrophic forgetting problem, in which the fine-tuned model tends to ``forget'' the source dataset's discriminative feature during fine-tuning~\cite{kirkpatrick2017overcoming}. To avoid this fallacy, we adopt the memory bank concept~\cite{wu2018memorybank} to enhance the model capacity by storing additional representations from the source dataset $\mathcal{D}_{s}$ in the memory bank. We compress knowledge as a prototype, a representative embedding for a group of semantically similar instances. This structure also ensures that shared knowledge no longer contains individual trade information. After pretraining, knowledge of frauds and non-frauds is clustered separately and condensed to generate transferrable prototypes. We apply \emph{K}-means clustering to extract the centroid set as prototypes for each class (i.e., $\mathcal{C}_{s} = \mathcal{C}_{\text{frauds}} \cup \mathcal{C}_{\text{non-frauds}}$). These extracted prototypes $\mathcal{C}_s$ are stored in the memory bank $M$, allowing knowledge transfer from the source country $s$. Recipients can utilize any subset of the transferred knowledge for their classification task and detect previously unrecognized illicit trades in $\mathcal{D}_{t}$.
 
\paragraph{Multi-Source Memory.} The memory bank $M$ can be expanded to adapt a multi-source scenario. For each source country, we pretrain the model independently using $\mathcal{L}_{SCL}$ and extract the prototype sets. The memory bank M can now contain an ensemble of multiple source countries' prototype sets as follows:
 
 \begin{equation}
    M = [\mathcal{C}_{s_1}, \mathcal{C}_{s_2}, ..., \mathcal{C}_{s_m}],
    \label{eq:multi_source}
\end{equation}

\noindent where $m$ is the number of source countries. %cardinality of source countries set.

% \begin{enumerate} 
% \item $\forall \mathbf{x}^{s} \in \mathcal{D}_{s}$, apply Supervised Contrastive Learning (SupCon)~\cite{khosla2020supervised} with their embedding $\mathbf{h_{x^s}}$. (SupCon을 통과해서 나온 결과물에 대한 notation 필요하면 넣기)

% \item Prototype을 만드는 방법: K-means clustering. SupCon 결과물의 L2-normalized 스페이스를 풀고 Euclidean space에서 \emph{k}-means 클러스터링을 돌려 프로토타입을 만듬. We extract prototypes $\mathbf{c}_s$ from source country $\mathcal{D}_{s}$. 이렇게 만들어진 프로토타입은 지원국에게 공유되어 지원국의 우범 선별 성능을 높임.

% \item 보강 설명: Fraud와 non-fraud의 차이 (discriminative feature)를 배우기 위해서는 Cross entropy보다는 SupCLR가 더 나은 방법임. 단순히 cross-entropy를 쓰면 discriminate한 feature를 구분하지 못함~\cite{khosla2020supervised}. 근데 SupCon의 경우에는, 같은 fraud끼리 모이게 하고, fraud와 non-fraud를 멀게 한다. 그래서 discriminative한 feature를 뽑기 쉬워진다. (Contrastive adaptation 같은 페이퍼도 씀) SupCLR의 결과물은 L2-normalized space에 저장된다. % Supervised contrastive learning (SupCLR)~\cite{}에서 영감을 받아, 도입국 

% \end{enumerate}
 
\subsection{Stage 2. Fine-Tune with Prototypes} 
% \sd{프로토타입만 전달해 주는 걸로 우리 모델을 확정하면 어떨지? Pretrained model $f$ 을 공개하는 것이 정책적으로 불가능할 수 있음. 또한, $f$를 함께 이용하여도 성능 증가폭이 크지 않았음.}

Given the memory bank  $M$ from $s$, the target country $t$ can fine-tune its fraud detection system $f(\cdot|\mathcal{D}_{t})$ using $\mathcal{D}_{t}$.
We design a fine-tuning step that is inspired by recent memory-based domain adaptation techniques that use enhancers for augmenting data representation~\cite{asghar2018progressive, liu2020open}. Following this concept, we let each target representation $\mathbf{h}_\mathbf{x}^t$ be 
refined with the representation $\mathbf{\bar{h}}_\mathbf{x}^{t|s}$ that is augmented with the source domain knowledge $\mathcal{C}_s$ in the memory $M$:
\begin{equation}
    \mathbf{\hat{h}}_\mathbf{x}^t = \mathbf{h}_\mathbf{x}^t + \mathbf{\bar{h}_x}^{t|s}.
    \label{eq:trasfer_hidden}
\end{equation}
The process of deriving $\mathbf{\bar{h}_x}^{t|s}$ is described in Eq.~7--10.

\paragraph{Source-Augmented Feature.}
Source-augmented feature $\mathbf{\bar{h}}_\mathbf{x}^{t|s}$ is computed via soft-attention toward the set of selected prototypes in $\mathcal{C}_{s}$~\cite{vaswani2017attention}.
Let k-th prototype be denoted $\mathbf{c}_{k}$. Each target transaction $\mathbf{h}_\mathbf{x}^t$ attends to prototype features via attention weights computed by dot product similarity:
\begin{align}
    w_{k}= \mathbf{\psi}( \mathbf{h}_\mathbf{x}^t \cdot \mathbf{c}_{k}) \\ 
    \mathbf{h}_\mathbf{x}^{t|s}= \sum_{k=1}^{|M|} w_{k}\mathbf{c}_{k},
    \label{eq:attention_weight}
\end{align}
where $\psi$(·) is a softmax function to normalize the dot product similarity scores across all prototypes.  %align이나 equation 전후로 space 안 주면 extra space 안 생깁니다, noindent 안 넣어도 됩니다. vspace 사용 지양

To regulate negative transfer between two domains~\cite{wang2019characterizing}, the network must calibrate how much knowledge to transfer from source to target. We calibrate source knowledge by using a single feed-forward network $g$ that computes additive attention weights between the source and target representations.
\begin{align}
    \mathbf{e_{x}} &= g(\mathbf{h}_\mathbf{x}^t, \mathbf{h}_\mathbf{x}^{t|s}) \\
    \mathbf{\bar{h}}_\mathbf{x}^{t|s} &= \phi_{r}([\mathbf{e_{x}} \odot \mathbf{h}_\mathbf{x}^{t|s}, \mathbf{h}_\mathbf{x}^t])
    \label{eq:calibrate}
\end{align}
where $\phi_{r}$ is a learnable parameter and $\odot$ is a Hadamard product.

\paragraph{Fine-Tuning with Target Data.} The target representation $\mathbf{\hat{h}}_\mathbf{x}^{t}$ is dynamically balanced between the direct transaction feature $\mathbf{h}_\mathbf{x}^{t}$ and the source-augmented feature $\mathbf{\bar{h}}_\mathbf{x}^{t|s}$ as in  Eq.~\eqref{eq:trasfer_hidden}. We fine-tune the target country's fraud detection system $f(\cdot|\mathcal{D}_{t})$:
\begin{align} 
    \mathcal{L}_{cls} = -\frac{1}{|\mathcal{D}_{t}|}\sum_{{\mathbf{x}, y}\in\mathcal{D}_{t}}  H(y, \hat{y})
    \label{eq:fine_tuning}
\end{align}
where $\hat{y} = f(\mathbf{\hat{h}}_\mathbf{x}^{t})$ is the predicted fraud score of transaction $\mathbf{x}$ and $H$ is a binary cross entropy.

% \begin{enumerate}

% \item Source-enhanced representation이란? 지원국 데이터 (source country, $\mathcal{D}_{s}$)를 함축하는 prototype vectors $\mathbf{c}_s$ 들을 활용하여, 정교화한 queried instance (우범 선별을 요하는 도입국의 transaction $\mathbf{x} \in \mathcal{D}_{t}$)의 representation

% \item Source-enhanced representation을 만드는 방법:
% % Memory bank $M$에 저장된
% 지원국 (source country)의 prototype $\mathbf{c}_s$들과 target country의 queried instance $\mathbf{x}$의 representation $\mathbf{h_x}$ 간의 attention weight sum

% \item 최종 representation: Queried instance의 최종 representation에 source-enhanced  representation를 추가 (element-wise sum). 

% \item 이렇게 합쳐진 representation을 이용하여 우범 여부를 선별 (cross-entropy loss)
% \end{enumerate}
\section{Results}

This section tests feasibility of the proposed knowledge sharing system in terms of detection performance, accommodating multiple sources, and dependency on the required log size and model components. 

\if 0
The evaluation was designed to answer the following questions. \mc{=rewrite below without mentioning DAS and discuss why these questions are important to customs officers=}
\begin{enumerate}
\item Domain Adaptation: How much does the performance of the fraud detection system improve when \ours{} is used?
\item Multi-source support: Can \ours{} be used when there are multiple donor countries?
\item Data size: How does the efficacy of \ours{} change depending on the amount of target data?
\item Component analysis: What are the best ways to utilize the \ours{} components?
\end{enumerate}
\fi

\subsection{Experimental Setting}

\subsubsection{Datasets}
\label{sec:experiments:settings:datasets}
We employed import declarations from four partner countries of WCO, whose data had been shared for the research purposes under the non-disclosure agreement. We refer to these countries M, C, N, and T in Table~\ref{tab:datastats}. The log size and GDP per capita vary by country, allowing us to test various source and target scenarios. Import declaration data is readily available by customs administrations of any country, and the data format is compatible with each other. We considered the common fields, including numeric variables on the item price, weight, and quantity and categorical variables on the HS6 code, importer ID, country code, and received office. The data also contained manual label information indicating whether each declaration was judged fraud or not, and the amount of tax raised after inspection. For the test purposes, these labels had been generated by inspecting all logs. We 
report the raw variables of import declarations in Table~\ref{Tab:variable_description}. 

%To generate this data, all speculative items had been inspected. The fraud inspection rate depends on the country. Some small countries are known to conduct 100\% manual inspection as of now. However, manual inspection will no longer be feasible as trade volume increases rapidly. In many developed countries, the inspection rate is around 1--5\% of all trade volume.  

%High-order categorical variables are flagged by quantifying their fraud history~\cite{kim2020date}. For example, importers whose fraud ranks are above 90\% were regarded as high-risk importers. Risky indicator features are marked as 1. 

\begin{table}[h!]
\centering
% \scriptsize
\small
\caption{Statistics of the datasets}
\label{tab:datastats}
\scalebox{1}{%
\begin{tabular}{ l | r  r  r  r } \toprule
    Country &  \textsf{M} & \textsf{C} & \textsf{N} & \textsf{T}  \\ \midrule
    % Periods & 2013--2016 & 2016--2019 & 2013--2017 & 2015--2019 \\ 
    Duration & 4 years & 4 years & 5 years & 5 years \\ 
    Num. imports & 0.42M & 1.90M & 1.93M & 4.17M  \\ 
    Num. importers & 41K & 9K & 165K & 133K  \\ 
    Num. tariff codes & 1.9K & 5.5K & 6.0K & 13.4K  \\ 
    % GDP per capita & \$300 & \$1,507 & \$2,230 & \$3,317 \\
    GDP per capita & \$300 & \$1,500 & \$2,200 & \$3,300 \\
    % Illicit rate &  1.64\% & 1.71\%   & 4.12\% & 8.16\%  \\ 
    \bottomrule
\end{tabular}
}
\end{table} 
\begin{table}[h!]
\centering
\caption{Overview of the transaction-level import data, in which the description of each variable are provided.}
\scalebox{0.75}{
\begin{tabular}{l|l}
\toprule
Variable & Description  \\ \midrule

\textit{Quantity} & The specified number of items\\
\textit{Gross weight} & The physical weight of the items \\
\textit{HS code}   &  Applicable item code based on the harmonised system \\
\textit{Country code}  & Country from which the goods were imported \\
\textit{CIF value} & Transaction value including the insurance and freight costs \\ 
\textit{Total taxes} & Tariffs calculated by initial declaration \\ \midrule
\textit{Illicit ($y$)} & Target binary variable indicating whether the item is fraud \\
\textit{Revenue} & Amount of tax raised after the inspection, used for evaluation \\
\bottomrule 
\end{tabular}}
\label{Tab:variable_description}
\end{table}

%The customs administrations in these four countries conduct manual inspections of all imported goods (i.e., achieving nearly 100\% inspection rate). Since the data obtained was under a complete inspection, illicitness of the transaction and charged tariffs are accurately labeled at the single-goods level. However, this practice is not sustainable, and they plan to reduce the physical inspection. Due to considerable trade volumes, the inspection rate in developed countries is known to be around 5\% or less. Therefore, the following experiments are designed considering these figures. \looseness=-1 

% \sd{경제 GDP 지수랑 데이터량이랑 비래함을 강조.}

\subsubsection{Evaluation Metric} Raised tax is one of the critical screening factors because customs import duties make up a substantial proportion of the tax revenue in many countries~\cite{Grigoriou2019WCO}. We hence consider the amount of tax raised from inspection as the performance indicator. In all experiments, we assume a 5\% inspection rate unless otherwise mentioned. We use the term Revenue@5\% to represent the ratio of the expected tax collected by inspecting 5\% of the trades sorted by the fraud detection algorithm out of the maximum revenue that could have been raised when the entire log was to be inspected.

\subsubsection{Training Details} 
The model $f$ is individually pretrained 10 epochs for every country and fine-tuned for 30 epochs. The sharpening temperature $\tau$ in Eq~\eqref{eq:SCL_loss} was set as 0.07 by following previous works~\cite{gunel2020supervised}. Smaller SCL loss temperatures benefit training more than higher ones, but extremely low temperatures are harder to train due to numerical instability. The number of prototypes was set to 500 per class, thus $|\mathcal{C}_s| = 1,000$. We vary this count later. We use the final month of each dataset as the test set, and the validation set was chosen as two weeks prior. The model was optimized using Ranger with a weight decay of 0.01~\cite{wright2021ranger21}. The batch size and learning rate were set as 128 and 0.005 for the entire training epochs. All models were run five times, and their average values were reported. 
% Other technical details are described in the Appendix.
% \footnote{\url{https://github.com/deu30303/RUC}} 

\subsection{Performance Evaluation}
%We examined the multi-source scenarios to compare our model against baselines and then investigated efficiency over varying labeled ratios.

The first set of experiments (Exp 1--2) aims to test the effectiveness of domain adaptation for single-source and multi-source scenarios. The next experiments (Exp 3--6) test the model's sensitivity and dependency to the target country's log size, model components, and memory-bank usage.

%The final experiments (Exp 6--7) demonstrate the effectiveness of the model with qualitative examples.

\subsubsection{Exp 1. Single-Source Scenarios} 

We inspect all combinations of the source and target pair as follows. The source country will share knowledge based on the 5\% of its fraud-like logs $\mathcal{D}_{s}$ sorted by $\hat{y}$. This knowledge is used in four different ways: (1) no sharing at all, (2) standard fine-tuning by sharing model parameters of $f(\cdot|\mathcal{D}_{s})$ directly with the target country, (3) transfer learning based on the adaptive knowledge consistency technique~\cite{abuduweili2021adaptive},
%to alleviate the negative transfer problem caused by discrepancy between the source and the target data =do we need to say this?=}, 
and (4) the proposed prototype-based memory bank. Below is a detailed explanation of the baselines used in our experiments.
\begin{itemize}
\item Target Only (No sharing): Measure the performance of a base fraud detection model $f(\cdot|\mathcal{D}_{t})$ trained on target data. The tree-based embedding model DATE is used~\cite{kim2020date} for $f$. DATE classifies and ranks illegal trade flows that contribute the
most to customs revenue when caught.
\item Vanilla Transfer: Sharing parameters of the fraud detection model $f(\cdot|\mathcal{D}_{s})$ from source country $s$ and directly fine-tune using target dataset $\mathcal{D}_{t}$. 
\item Adaptive Transfer: Adaptive Knowledge Consistency (AKC) technique~\cite{abuduweili2021adaptive} is used to cope with the risk of negative transfer caused by the discrepancy between the source and target countries. It constrains the mean square error between the pre-trained feature extractor outputs and the target feature extractor outputs by adaptively sampling the target dataset $\mathcal{D}_{t}$. We perform a hard filter according to the sample importance by sorting fraud score difference between the source and target pre-trained models. The features with a lower 20 \% score difference were selected. 
\end{itemize}

At the receiving end, we assume the target country has a weaker infrastructure and utilizes only 1\% of trades for training. This assumption is simulated by randomly masking the log labels. The base fraud detection model $f$ uses implementations of the tree-based embedding model~\cite{kim2020date}, whose performance is reported in the column indicated as `Target Only' in Table~\ref{tab:singlesource}. The Revenue@5\% performance for the target country M is the lowest, which may be contributed by the smallest log size. Note that country M has the lowest GDP per capita in Table 1. The efficacy of the detection model increases when utilizing knowledge shared by countries with larger log sizes (i.e., choosing Country T as the source country), although with some exceptions. Since the illicit rates of these countries are different, we only compare increments in performance within each target country.

\begin{table}[h!]
\centering
% \scriptsize
\caption{Revenue@5\% performance under four different settings: (1) when no knowledge is shared and only the target country's logs are used for detection, (2) when the source country's knowledge is shared via direct parameter sharing, (3) when knowledge is shared via transfer learning to mitigate the negative transfer problem, and (4) when knowledge is further embedded as prototypes and shared as proposed in our system. The ratio of tax raised by \ours{} is substantial compared to other settings.
%Customs selection performance in target country boosted significantly by using the domain adaptation technique \ours{}. For this experiment, small target dataset is used (1\% labeled).
} % \sd{Random baseline의 Revenue@5\% 성능이 0.05 인데, DATE (M)의 성능과, Fine-tune (N-C)의 성능이 너무 낮은 것 같아요.}
% 1) 비슷한 국가를 source로 활용하였을 때 성능이 높아졌다는 점 어필 가능. 
% 2) T 데이터가 가장 잘 사는 나라이고 데이터 양이 많아서, 이를 source로 활용하였을 때 성능이 일반적으로 더 높아짐. 
% 3) 반면, 함께 source와 target 데이터를 함께 DATE의 input으로 넣었을 때는 Negative transfer 등의 요인으로 T 데이터를 함께 썼을 때 성능 증가가 오히려 적었다.
\label{tab:singlesource}
% \resizebox{\linewidth}{!}{%
\small
\begin{tabular}{l|cccc} \toprule
    Test & Target & Vanilla & Adaptive  &  Proposed \\
    Setup & Only & Transfer & Transfer &  \ours{} \\ \midrule
    \textsf{N} $\rightarrow$ \textsf{M} & 0.0466 & 0.0681&  0.1091 & 0.5152 \\
    \textsf{C} $\rightarrow$ \textsf{M} & & 0.0869 & 0.1405 & 0.5271 \\
    \textsf{T} $\rightarrow$ \textsf{M} &  & 0.2112 & 0.2458 & 0.2966 \\
    % \textsf{N}$+$\textsf{T} $\rightarrow$ \textsf{M} & & & & 0.5107\\
    % \textsf{N}$+$\textsf{C} $\rightarrow$ \textsf{M} & & & & 0.5385\\
    % \textsf{C}$+$\textsf{T} $\rightarrow$ \textsf{M} & & & & 0.5426\\
    % \textsf{N}$+$\textsf{C}$+$\textsf{T} $\rightarrow$ \textsf{M} & & & & 0.5435\\
    \midrule
    \textsf{M} $\rightarrow$ \textsf{C} & 0.0853 & 0.1806  & 0.1843 & 0.1915\\
    \textsf{N} $\rightarrow$ \textsf{C} & & 0.0297 & 0.1081 & 0.1794\\
    \textsf{T} $\rightarrow$ \textsf{C} & & 0.0719 & 0.1402 &  0.2271\\
    % \textsf{M}$+$\textsf{N} $\rightarrow$ \textsf{C} & & & & 0.2154\\
    % \textsf{M}$+$\textsf{T} $\rightarrow$ \textsf{C} & & & & 0.2339\\
    % \textsf{N}$+$\textsf{T} $\rightarrow$ \textsf{C} & & & & 0.2328\\
    % \textsf{M}$+$\textsf{N}$+$\textsf{T} $\rightarrow$ \textsf{C} & & & & 0.2340\\  
    \midrule
    \textsf{M} $\rightarrow$ \textsf{N} & 0.1837 & 0.1018 & 0.3244 & 0.6681\\
    \textsf{C} $\rightarrow$ \textsf{N} & & 0.2368 & 0.4043 & 0.6863 \\
    \textsf{T} $\rightarrow$ \textsf{N} & &  0.1119 & 0.3032 & 0.7286 \\
    % \textsf{M}$+$\textsf{C} $\rightarrow$ \textsf{N} & & & & 0.6887\\
    % \textsf{M}$+$\textsf{T} $\rightarrow$ \textsf{N} & & & & 0.7387\\
    % \textsf{T}$+$\textsf{C} $\rightarrow$ \textsf{N} & & & & 0.7347\\
    % \textsf{M}$+$\textsf{C}$+$\textsf{T} $\rightarrow$ \textsf{N} & & & & 0.7395\\ 
    \midrule
    \textsf{M} $\rightarrow$ \textsf{T} & 0.1541 & 0.3005 & 0.3037 & 0.3181 \\
    \textsf{C} $\rightarrow$ \textsf{T} & & 0.2844  & 0.3012 & 0.3073\\
    \textsf{N} $\rightarrow$ \textsf{T} & & 0.2007 & 0.3033 & 0.3282\\
    % \textsf{M}$+$\textsf{C} $\rightarrow$ \textsf{T} & & & & 0.3161\\
    % \textsf{M}$+$\textsf{N} $\rightarrow$ \textsf{T} & & & & 0.3292\\
    % \textsf{C}$+$\textsf{N} $\rightarrow$ \textsf{T} & & & & 0.3291\\
    % \textsf{M}$+$\textsf{C}$+$\textsf{N} $\rightarrow$ \textsf{T} & & & & 0.3286\\ 
    \bottomrule
    \multicolumn{5}{l}{\textsf{S} $\rightarrow$ \textsf{T} indicates the direction of knowledge flow from} \\
    \multicolumn{5}{l}{the source  country \textsf{S} to the target country \textsf{T}.}
    % }
\end{tabular}
\end{table}

When knowledge is shared in any form, the tax revenue from fraud detection generally increases. For the target country M, the use of \ours{} from country N increases the tax revenue 11 times from 0.0466 to 0.5152. The exact benefit differs by country, implying that various factors like the log similarity between countries may play a role. Parameter sharing (i.e., column indicated as Vanilla Transfer) sometimes leads to degraded performance due to negative transfer. Transfer learning via the adaptive knowledge consistency technique (i.e., Adaptive Transfer) no longer shows this limitation. Sharing knowledge in the form of prototypes shows the best performance in terms of tax raised, as indicated in the final column of the table. Note this method also increases data protection of the source countries.

\subsubsection{Exp 2. Multi-Source Scenarios} 

\begin{figure}[h!]
    \centerline{
    \includegraphics[width=0.95\columnwidth]{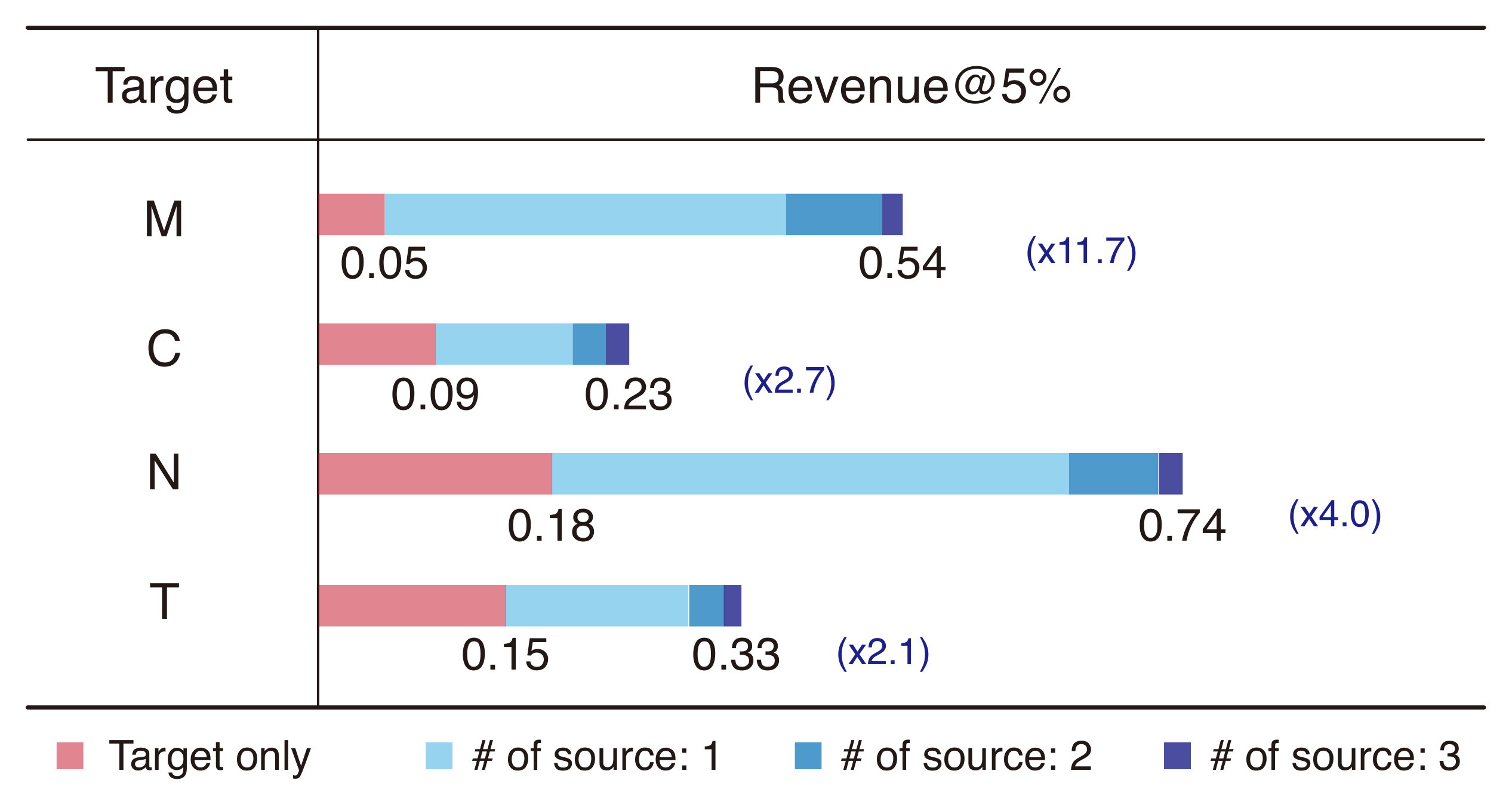}}
    \caption{Collected duties are expected to increase 2.1--11.7 times for all tested countries when shared knowledge contributed by multiple sources is used (i.e., blue bars) compared to relying on local knowledge alone (i.e., red bars).
    %Revenue@5\% reports the ratio of tax raised by inspected the 5\% of the most likely fraud declarations. The score of 1.0 represents the total tax raised when 100\% of the declarations had been inspected. 
    %Compared to using the local knowledge (i.e., red bar), shared knowledge increases the tax revenue from fraud detection (i.e., blue bars).
    %The increment is 10 times for country \textsf{M} with the smallest log size, but also substantial (i.e., 2 times) for country \textsf{T} with the largest log size.
    }
    \label{fig:multisource}
\end{figure}

Accumulating knowledge from more than one country further increases the performance of the fraud detection system. Figure~\ref{fig:multisource} shows the average Revenue@5\% for using no source at all, a single source, two sources, and three sources. To compute, we averaged the performance of all possible combinations according to the number of source countries. The additional gain in raised tax is most noticeable for using a single source, yet there is a revenue gain with every added source. When all available source knowledge is used, the target countries benefit nearly 2 to 11 times increase in total tax raised. \looseness=-1

%Pulling knowledge from more than one country can further boost the performance of the fraud detection system, as one might expect. Figure~\ref{fig:multi} shows the relative revenue of inspecting 5\% of the most fraud likely declarations when using logs from the target country alone to using the shared knowledge from an increasing number of source countries. A considerable benefit is shown when any source information is used, yet we also observe a gradual increase in benefits with added sources.

%We reconstruct the setting where the WCO maintains an expandable memory module by receiving multiple country inputs and maximizing their support to target countries. \ours{} performance using a multi-source memory module. To illustrate, we averaged the performance of all possible combinations according to the number of source countries. The performance of a model using multi-source memory tends to be better than the single-source case. As the number of source countries increases, the target performance gradually increased. Through this result, projects to build a regional the fraud detection system will gain momentum.

%\subsubsection{Exp 3. Efficacy According to Target Size}
\subsubsection{Exp 3. Dependency on Log Size} 

\begin{figure}[h!]
    \centerline{
    \includegraphics[width=1.00\columnwidth]{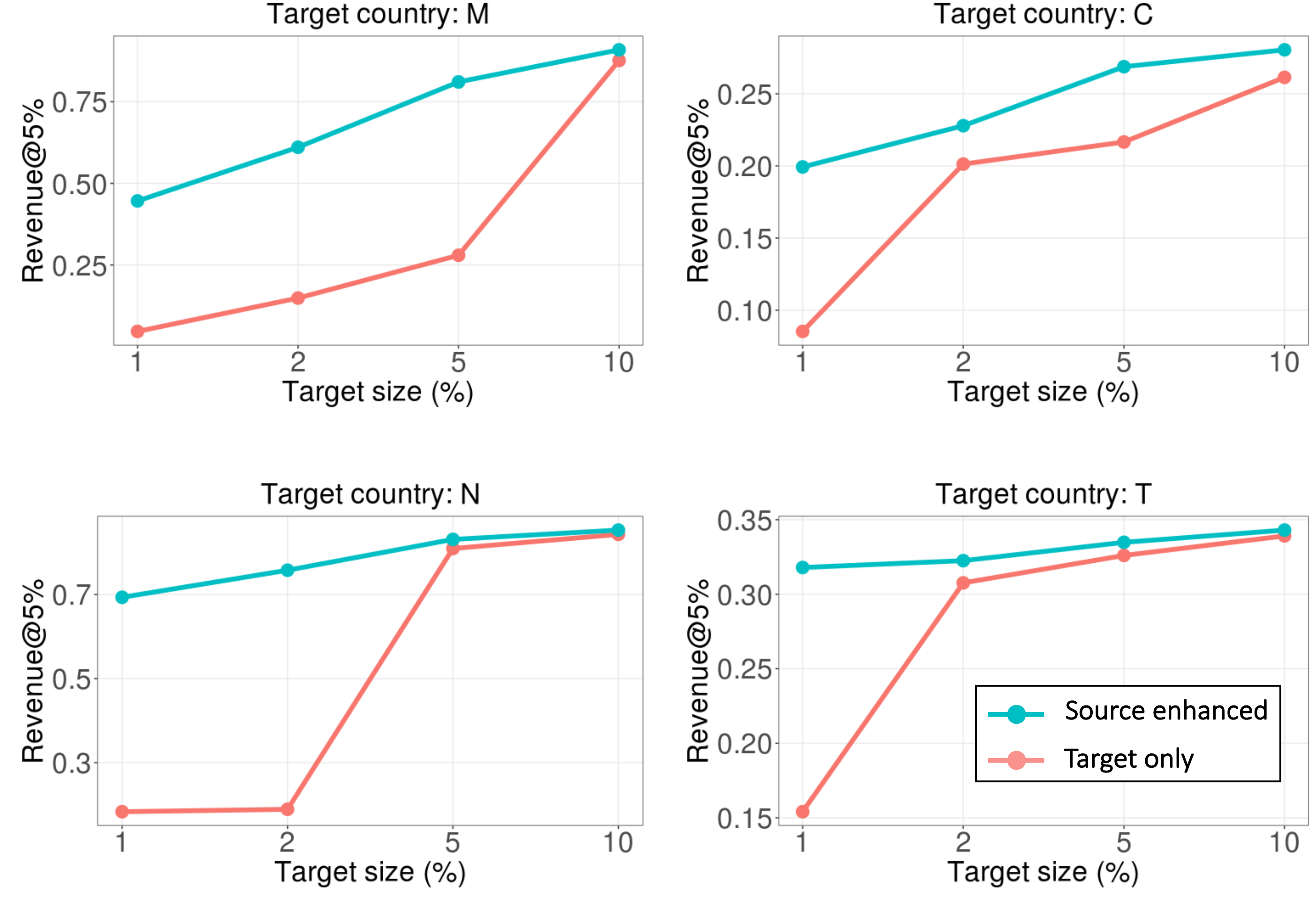}}
    \caption{Performance improvement as the log size of the target country increases. The shared knowledge brings the most considerable benefit when the available log size is the smallest (i.e., 1\%) and for countries with the weaker economy (i.e., country M).} 
    \label{fig:target_size}
\end{figure}

Next, we examine measurable effects such as the log size. Figure~\ref{fig:target_size} shows the average benefit of utilizing a single source versus no shared knowledge over the increasing ratio of log size for each target country. The performance gap is the largest when the log size is 1\%, implying \ours{} can benefit countries with weak infrastructure the most. Yet, the continued benefit is observed for increasing log size from 2\% to 10\% in the target country, even at a marginal level. The benefit for country T is the smallest, which has the highest GDP per capita. Nonetheless, we emphasize that sharing knowledge increases the capacity to manage risk in the neighboring participating countries, ultimately curtailing illicit trades in that region. \looseness=-1

%is the most helpful for countries with the smallest label set, yet it also helps improve detection for countries that already have collected sizable patterns of frauds. \mc{=Is the figure reporting average values or the best values?=}

%The shared knowledge of fraud can benefit target countries with poor infrastructure and the source countries with a greater degree of labeling. 

%To show that \ours{} is effective even when the target country's data set has been established to some extent, we experiment by gradually increasing the size of the target data. We created target datasets of different sizes. As seen in Figure~\ref{fig:target_size}, \ours{} is the most effective when the size of target data is the smallest. As the target dataset increases, the performance of the base model increases, but there is still a performance improvement by using \ours{} with source data. 
% \sd{Figure 4가 담고 있는 정보가 너무 적어서 실험을 보강했으면 해요. Target data를 1\%, 2\%, 5\%, 10\% 이런 식으로 바꿔가며 \ours{} 성능폭이 어떻게 변해가는 지 보여주면 좋겠어요.}

% \input{pgfplots/target_size}
% 현재 Country N 데이터 (100\% labeled, source)를 활용해, Country M 데이터 (5\% labeled, target)의 예측 성능을 높였고, 표~\ref{tab:results_main}에서 유의미한 결과를 확인할 수 있다.

% \input{pgfplots/target_size}

% \subsection{Component Analysis}
%For evaluation,  we first compared the performance of our model against other baselines. Then, we examined the multi-source situation to verify our prosed model’s scalability. Lastly, we investigated the proposed approach’s efficiency according to the target labeled ratio.

\subsubsection{Exp 4. Dependency on Model Components} 
%So far, we showed that \ours{} improves the fraud detection system's performance substantially under multiple situations. 
% \mc{=Shortened=}
%We now examine how much \ours{} components contribute to the performance gain via ablation study and parameter analysis. \ours{} utilizes diverse transferable learning components to augment with source information, so we remove each component from the full model to assess its efficacy. 
Table~\ref{tab:Ablation} reports the performance degradation when each component of the model is missing. The experiments are done with the target country N with a 1\% labeled setting. The full model with all components performs the best, justifying the role of each component. Mainly, domain invariant encoding of \ours{} contributes to the model performance the most. This experiment considers five scenarios.

\begin{itemize}

\item  Full model: \ours{} containing all components.

\item  Without domain invariant encoding: The model without Eq.~\ref{eq:outer_product}--\ref{eq:final_hidden} that simply concatenate between HS6 code and transaction embedding. 

\item  Without contrastive learning: The model without Eq.~\ref{eq:SCL_loss}, which uses cross-entropy loss instead of SCL loss for pretraining source dataset. 

\item  Without memory bank: The model without Eq.~\ref{eq:attention_weight}, which fine-tunes the target data without utilizing the information of the source memory bank. 

\item  Without domain calibration: The model without Eq.~\ref{eq:calibrate} that only considers the interaction between source feature and target feature.

\end{itemize}
\begin{table}[h!]
\centering
\caption{Ablation results on country N}
\scalebox{0.9}{
\begin{tabular}{l|ccc}
\toprule
Setup & \textsf{M} $\rightarrow$ \textsf{N} & \textsf{C} $\rightarrow$ \textsf{N}        & \textsf{T} $\rightarrow$ \textsf{N}     \\ \midrule
Full model      & \textbf{0.6681}  & \textbf{0.6863}  & \textbf{0.7286} \\
Without domain invariant encoding           &     0.4979      & 0.6668 & 0.2101\\
Without contrastive learning      &      0.5149              & 0.6766 & 0.7018\\
Without memory bank &      0.6667              & 0.6564 & 0.6762\\
Without domain calibration          &     0.6670      & 0.6657 & 0.6825\\

\bottomrule 
\end{tabular}
}
\label{tab:Ablation}
\end{table}

% \begin{table}[!h]
% \centering
% \scalebox{1.0}{\begin{tabular}{l|ccc}
% \toprule
% Setup & \textsf{M} $\rightarrow$ \textsf{N} & \textsf{C} $\rightarrow$ \textsf{N}        & \textsf{C} $\rightarrow$ \textsf{N}     \\ \midrule
% with all components      & 0.6651  & 0.6863  & 0.7286 \\
% without contrastive learning      &      0.6813           & 0.6960 & 0.6924\\  
% \bottomrule 
% \end{tabular}}
% \caption{Ablation results on country N dataset}
% \label{Tab: Ablation}
% \end{table}

\subsubsection{Exp 5. Efficacy of Memory Contents} 
Additionally, we discuss the augmentation effect when the source knowledge is transferred through the memory bank. For comparison, we consider a baseline of random memory bank with the same size, inspired by the prior work that randomly injected noise provides practical benefits to deep models~\cite{poole2014analyzing}. 
Table~\ref{tab:compare_augment} shows that memory bank usage provides non-trivial improvements than using random memory banks. This finding confirms that information shared by source countries is successfully leveraged in the form of prototypes.
\begin{table}[h!]
\centering
\caption{Efficacy of memory-based augmentation}
\scalebox{0.85}{\begin{tabular}{l|cccc}
\toprule
 & \multicolumn{4}{c}{Target country}     \\ \cmidrule{2-5}
Memory setup & \textsf{M} & \textsf{C} & \textsf{N} & \textsf{T}     \\ \midrule
No memory     &   0.5183 &   0.2233 & 0.6762 & 0.3190\\
Random     &      0.5241  & 0.2265 & 0.7126 & 0.3238\\
Three countries  & \textbf{0.5435}  & \textbf{0.2340}  & \textbf{0.7395} & \textbf{0.3286}\\
\bottomrule 
\multicolumn{5}{l}{A pretrained network $f(\cdot|\mathcal{D}_{s})$ from Eq.~\eqref{eq:SCL_loss} is shared together} \\
\multicolumn{5}{l}{regardless of the memory bank setup.}
\end{tabular}}
\label{tab:compare_augment}
\end{table}

\subsubsection{Exp 6. Dependency on Prototype Count}
Knowledge of frauds is shared as compressed representations called prototypes. The number of prototypes can determine the resolution of transferred knowledge. For instance, one may assume selective fraud patterns may be shared when limiting the prototype count. We test the sensitivity of the model on prototype count $|\mathcal{C}_s|$ by setting it from 10 to as large as the source data size $|\mathcal{D}_{s}|$. Table~\ref{tab:proto} shows detection performance remains intact by $|\mathcal{C}_s|$, implying that the model is not sensitive to the choice of hyper-parameters as long as the prototypes are used. 
\begin{table}[h!]
\centering
\caption{Sensitivity analysis to the number of prototypes for country N. $|\mathcal{D}_{s}|$ implies that the entire embeddings are transferred instead of the clustered prototypes.}
\small
\scalebox{1}{
\begin{tabular}{c|ccccc}
\toprule
$|\mathcal{C}_s|$ & 0 & 10 & 100        & 1,000   &   $|\mathcal{D}_{s}|$ \\ \midrule
\textsf{M} $\rightarrow$ \textsf{N} & 0.6667  & 0.6714  & 0.6721  & 0.6681 & 0.6709\\
\textsf{C} $\rightarrow$ \textsf{N} & 0.6564 & 0.6656      & 0.6727 & 0.6863 & 0.6711\\
\textsf{T} $\rightarrow$ \textsf{N} & 0.6762 & 0.7379      & 0.7307 & 0.7286 & 0.7314\\
\bottomrule 
\end{tabular}
}
\label{tab:proto}
\end{table}

% Best performing pretrained network $f(\cdot|\mathcal{D}_{s})$ is shared for fair evaluation. 

% dropping any component results in performance degradation.

\section{Discussion and Conclusion}

This paper presented a first-of-a-kind knowledge sharing system for multi-national customs administrations. Below we discuss the implications of findings in terms of risk management and protection of trade information.

\subsection{Discussion}

\ours{} has shown substantial revenue potential by utilizing a relatively small fraction of fraud-like logs. Custom offices with weaker infrastructure will likely observe the largest increase in raised tax, whereas the more equipped customs offices will see a smaller gain. However, the benefit needs to be interpreted in two other aspects. First, once the system is set up, the cost of fine-tuning for additional prototypes will be minimal. Hence, any increase in detection performance translates to a potentially substantial additional tax revenue in the target country. Second, empowering countries with weak infrastructure will enable the detection of new fraud patterns that were previously unseen. The new fraud patterns collected from target countries will be shared back to the memory bank. This helps strengthen policy against illicit trades by removing weak spots in the global trade chain and further benefit participating countries~\cite{wang2018wcj}.

%\subsubsection{Examples of Newly Detected Frauds}  

\begin{figure}[t!]
    \centerline{
    \includegraphics[width=1\columnwidth]{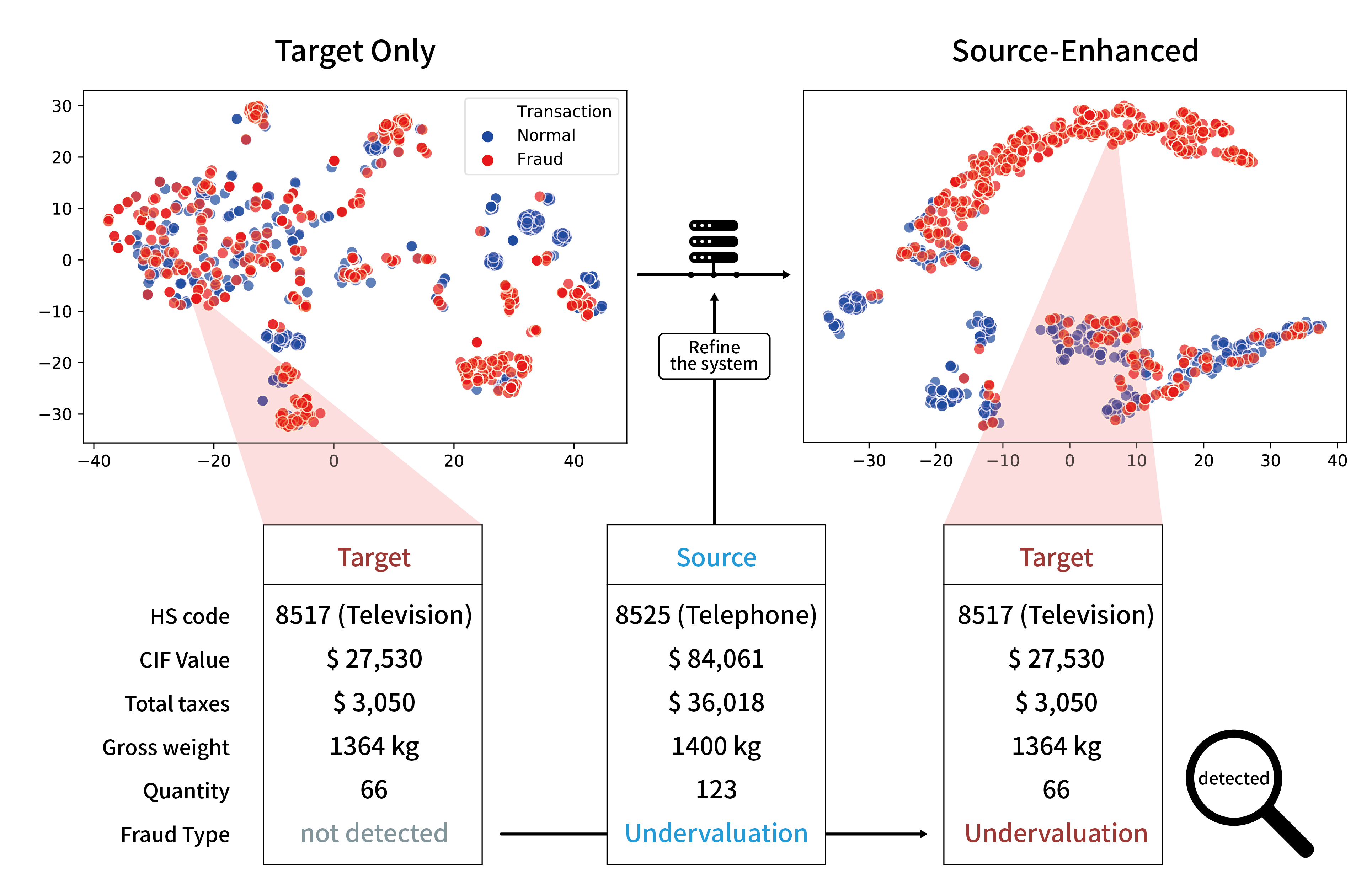}}
    \caption{t-SNE plots of the learned embeddings (T $\rightarrow$ N), when the model is trained only with target-only feature (left) and with source enhanced feature using \ours{} (right). Fraud cases are successfully detected after receiving prototypes from the source country. Similar fraud examples from the source country help flag this case. }
    \vspace{-1mm}
    \label{fig:qual}
\end{figure}

Qualitative analysis can be used to show how well the shared information discriminates against illicit transactions. Figure~\ref{fig:qual} compares embedding results of the target country, without knowledge (i.e., left) and with knowledge (i.e., right). Using the source-enhanced features helps better distinguish frauds from non-frauds. This particular example shows a fraud case that was newly detected by \ours{}. Without the shared knowledge, this declaration would have been missed. In contrast, the knowledge augmentation placed this declaration more closer to the fraudulent cluster in the embedding. \looseness=-1

The proposed method of sharing prototypes is a safe way to transfer knowledge across heterogeneous administrative domains. To our understanding, it is nearly impossible to re-identify information from a memory bank. Here, we show the input data utilized for knowledge building is statistically different from the vast ``normal'' trades that include critical information about trade partners and prices---information that is confidential and of concern in data sharing. 

\begin{figure}[t!]
    \centerline{
    \includegraphics[width=1\columnwidth]{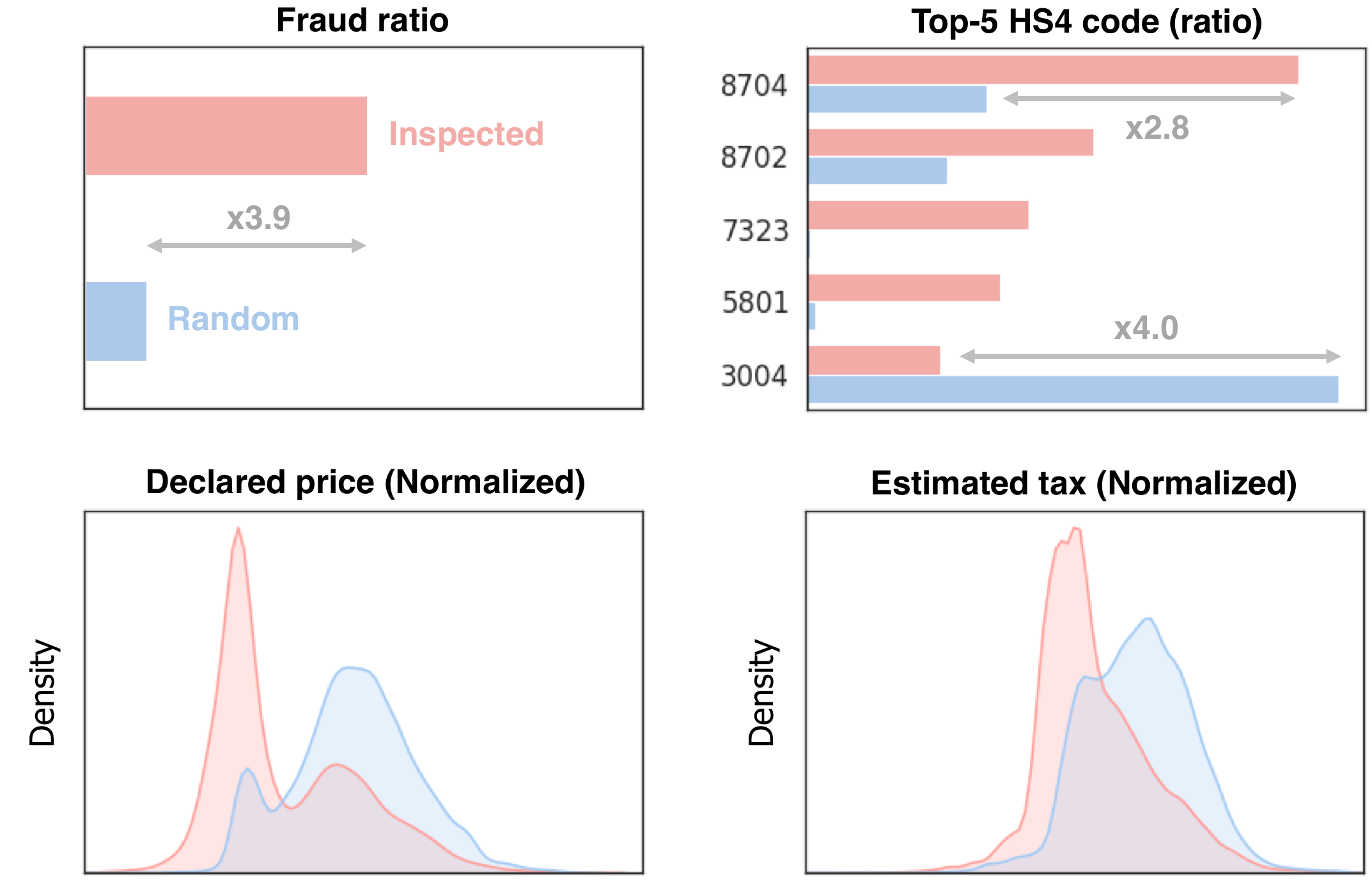}}
    \caption{Shared knowledge of frauds is distinct from the mass of the normal trade volume contributed by member offices. Exact figures are hidden.}
    \label{fig:difference}
\end{figure}

Figure~\ref{fig:difference} highlights some of the differences observed for fraud-suspected logs and normal logs. The logs utilized by the model have nearly four times higher fraud ratio than normal logs. In addition, the product categories that frequently appear in fraud-like logs and normal logs have different rank orders, as illustrated in the examples. The average declared price (CIF value) indicates that the initial reporting tariffs written in the declaration form are substantially lower for the inspected logs. Only fraud-suspected logs are used for knowledge building, and the normal data are excluded in the data embedding step.
%, so customs administration can alleviate concerns about sensitive information being exposed.

% difference, and the  
% The fraud ratio, the distribution of traded product categories (HS4 code), the declared price ranges (CIF value) and the corresponding initial tariffs have little overlap. 
% Only the fraud-suspected logs are used for knowledge building and the normal data are entirely excluded in the data embedding step.

% \sd{
% Since the inspected transactions are selected by the fraud detection model, we observed that the fraud rate of inspected transactions is four times higher than that of overall trades. Accordingly, average declared price and estimated initial tax of the inspected transactions were much low. In addition, HS code distribution between the two dataset is different, showing that the speculative source data used for knowledge sharing is inherently different from overall trade pattern.
% }

% \begin{figure}[h!]
%     \begin{subfigure}[b]{.49\linewidth}
%         \centering\captionsetup{width=.95\linewidth}%
%         \includegraphics[width=\linewidth]{example-image-golden}
%         \caption{Fraud rate}
%     \end{subfigure}
%     \begin{subfigure}[b]{.49\linewidth}
%         \centering\captionsetup{width=.95\linewidth}%
%         \includegraphics[width=\linewidth]{figures/HS_distribution.png}
%         \caption{Top-5 HS code}
%     \end{subfigure}
    
%     \begin{subfigure}[b]{.49\linewidth}
%         \centering\captionsetup{width=.95\linewidth}%
%         \includegraphics[width=\linewidth]{figures/CIF_distribution.png}
%         \caption{Declared price}
%     \end{subfigure}
%     \begin{subfigure}[b]{.49\linewidth}
%         \centering\captionsetup{width=.95\linewidth}%
%         \includegraphics[width=\linewidth]{figures/TAX_distribution.png}
%         \caption{Estimated tax}
%     \end{subfigure}
%     \caption{Source country's fraud-like data $\mathcal{D}_s$ and its overall trades are essentially different.}
%     \label{fig:difference}
% \end{figure}

\subsection{Concluding Remark}
There is an increasing need to share knowledge in data-critical sectors, including customs, medicine, finance, and science. We tested the premise that shared knowledge will facilitate and advance risk management in customs fraud detection. Domain adaptation techniques are one way to build collective knowledge across international administrations. Testing with million-scale data demonstrated that transferable knowledge increases the detection performance for all participating countries. A centrally managed memory bank system that we explored is one such possibility.  

As global trade sees dynamic changes due to events like COVID-19, maintaining an agile risk management system has become ever more important. Strengthening the data science capacity at customs administrations and collaboration will be critical as illicit traders continue to use vantage points in the trade network. Starting from this proof of concept, we envision more countries agreeing on the legal support for sharing prototypes outside their administrative domains. When more countries participate in the regional memory bank platform, we expect knowledge sharing will better handle the data challenges. Furthermore, the regional ties will help enable positive trade agreements among the participating countries.

%% \sd{Removed by the request of WCO.}
%Starting from this proof of concept, discussions are underway to plan an active service at the WCO. The memory bank system requires that the source countries agree on the legal support for sharing prototypes outside their administrative domains. We envision this will be enabled via a central cloud service managed by WCO that will facilitate knowledge sharing while safeguarding data. 

%Increasing data science capacity within customs administrations is crucial to introduce technological advancements with less central support. As more countries participate in the platform, we expect to see which combination of countries the knowledge sharing will work well or whether the methodology is robust in situations where shared knowledge is imperfect.

%\sd{=자명한 얘기라 코멘트했어요=}
% The transferable components help avoid negative transfer and catastrophic forgetting between the source and target countries, achieving the best performance against state-of-the-art fraud detection models.

% More participation from members countries in the form of providing useful knowledge as well as receiving suitable knowledge.

% \section{Acknowledgement}
\section*{Acknowledgment}
This work was supported by the Institute for Basic Science (IBS-R029-C2, IBS-R029-Y4) and the National Research Foundation (No. NRF-2017R1E 104 1A1A01076400) funded by the Ministry of Science and ICT in Korea. We thank the World Customs Organization (WCO), the Capacity Building Directorate, and the partner countries for support in data access. The views and conclusions contained herein are those of the authors and should not be interpreted as necessarily representing the official policies or endorsements, either expressed or implied, of WCO.

% Use \bibliography{yourbibfile} instead or the References section will not appear in your paper
% \balance
\bibliography{aaai22}
% \nobibliography{aaai22}

% \input{6_supplements.tex}
% \input{7_rebuttal}
% \bibliography{aaai22}

\end{document}